\documentclass[letterpaper,10pt,journal,twoside]{IEEEtran}

\usepackage{cite}
\usepackage{bm}
\usepackage{amsmath}
\usepackage{amssymb}
\usepackage{graphicx}
\usepackage{hyperref}
\usepackage{booktabs}
\usepackage{subcaption}
\usepackage{multirow}
\usepackage{diagbox}

\begin{document}

\title{WAND: Learning Robust Navigation under Complex Wind Disturbances and Dense Obstacles for Quadrotors}

\author{Zhonghan~Tang$^{1,2}$, Chenhui~Li$^{2}$, Shuai~Liang$^{2}$,
Zhongrui~You$^{2}$, Jianan~Li$^{2}$, Bin~Zhao$^{2,3}$,
Zhigang~Wang$^{2}$, and Xuelong~Li$^{4}$%
\thanks{This work was supported by Shanghai Artificial Intelligence Laboratory. \emph{(Zhonghan Tang and Chenhui Li contributed equally to this work.)} \emph{(Corresponding author: Zhigang Wang.)}}%
\thanks{$^{1}$Zhonghan Tang is with the Department of Electronic Engineering
and Information Science, University of Science and Technology of China,
Hefei 230027, China (e-mail: tzh123@mail.ustc.edu.cn).
$^{2}$Zhonghan Tang, Chenhui Li, Shuai Liang, Zhongrui You, Jianan Li,
Bin Zhao, and Zhigang Wang are with Shanghai Artificial Intelligence Laboratory,
Shanghai 200232, China (e-mail: tangzhonghan@pjlab.org.cn; wangzhigang@pjlab.org.cn).
$^{3}$Bin Zhao is also with Northwestern Polytechnical University,
Xi'an 710072, China (e-mail: bin@nwpu.edu.cn).
$^{4}$Xuelong Li is with the Institute of Artificial Intelligence
(TeleAI), China Telecom, Beijing 100033, China
(e-mail: xuelong\_li@ieee.org).}
}

\maketitle

\begin{abstract}

Robust navigation in cluttered environments remains a fundamental challenge for quadrotors, particularly when strong wind disturbances arise, which perturb vehicle dynamics, limit control authority, and substantially increase collision risk. Existing learning-based navigation policies typically rely on obstacle perception and proprioceptive observations, requiring the policy to infer time-varying disturbance effects implicitly and thereby limiting robustness under partial observability. This paper proposes WAND (Wind-Aware Navigation with Disturbance Estimation), a reinforcement learning framework for navigation under time-varying wind disturbances in dense obstacle fields. Specifically, WAND estimates wind-induced disturbance acceleration from historical proprioceptive states using a Temporal Convolutional Network (TCN). This estimation is integrated into the policy via a zero-initialized residual module, \emph{WindAdapter}, while simultaneously providing feedforward compensation for low-level control. The dual use of the estimate couples disturbance-conditioned navigation with feedforward disturbance rejection. Across 12 wind-disturbed simulation settings, WAND improved the observed success rate by 8.3 percentage points on average relative to feedforward compensation alone. Controlled opposite-crosswind experiments further showed wind-direction-dependent trajectory adaptation. In indoor fan-induced flight tests, WAND succeeded in 18 of 20 trials, demonstrating the feasibility of real-time onboard navigation.

\end{abstract}



\section{INTRODUCTION}

\IEEEPARstart{I}{n} recent years, micro quadrotors have seen rapid adoption in exploration \cite{2021epic}, emergency rescue \cite{rescue2022},  and urban delivery \cite{delivery2017}. Driven by breakthroughs in low-power computing and micro-sensing \cite{lowcost_diff2025, lowcost_rgb2025}, these vehicles can now operate under extreme Size, Weight, and Power (SWaP) constraints, granting them access to dense and confined environments. In such settings, the wind disturbance experienced by the vehicle can vary over time as the airflow interacts with surrounding structures, for example, through flow acceleration near gaps or recirculation around obstacles \cite{UCE2006, edge_flow2008}. Due to their limited actuation authority, micro quadrotors are highly susceptible to unsteady airflow, increasing the risk of collision in cluttered environments. 

Traditional navigation stacks separate geometric planning from low-level control. Geometry-oriented planners \cite{fast_planner2019,ego_planner2021,super2025} focus on trajectory feasibility, whereas robust planners \cite{tube_mpc2017,motion_plan2021} and predictive controllers \cite{cmpcc2021} account for disturbances or tracking uncertainty but depend on assumed models and sufficient control authority. Strong, rapidly varying wind can therefore leave residual tracking errors that erode trajectory clearance in confined environments.

\begin{figure}[t] 
    \centering
    \includegraphics[width=\linewidth]{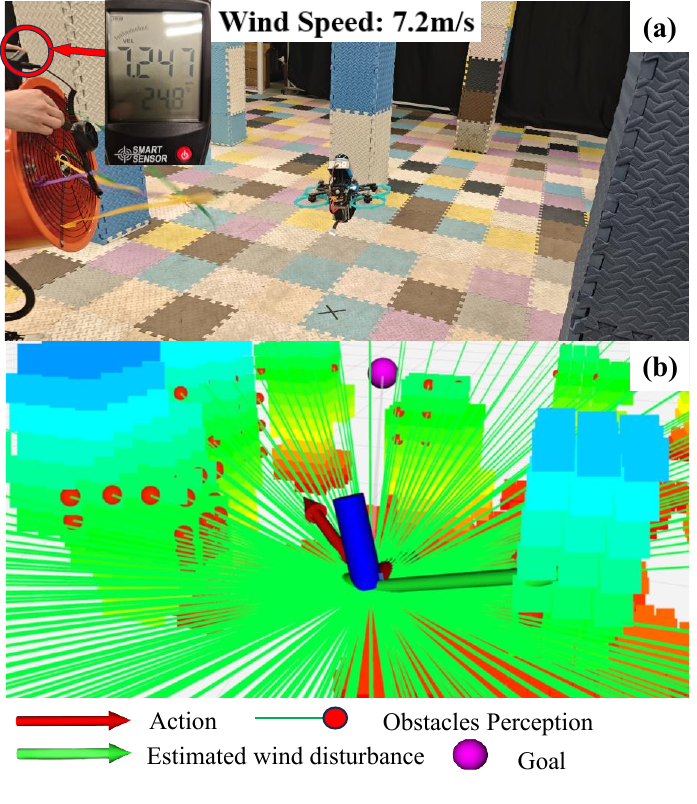}
    \caption{Real-world deployment of WAND. (a) A quadrotor navigates a
    cluttered indoor scene under fan-induced airflow; the inset shows an
    anemometer reading of approximately $7.2\,\mathrm{m/s}$. (b)
    The corresponding visualization in RViz.}
    \label{flight example}
\end{figure}

In contrast, learning-based methods integrate perception and control through learned navigation policies. NavRL \cite{navrl2025} learns safe navigation in dynamic clutter, while end-to-end approaches map LiDAR \cite{cloud2025,e2e_learning2025} or visual features \cite{learning_agile2023} to low-level commands and support sim-to-real transfer \cite{ctbr2022}. However, these policies primarily condition on obstacle perception and proprioception rather than explicit wind estimates. They therefore infer time-varying aerodynamic effects implicitly, although the same observation can require different actions under different winds. This partial observability complicates robust wind-aware navigation.

To address this issue, we propose WAND, a wind-aware RL framework for safe navigation in windy, cluttered environments. A Temporal Convolutional Network (TCN) estimates wind-induced disturbance acceleration from proprioceptive histories, and the proposed \emph{WindAdapter} encodes this estimate into wind-conditioned features. The policy combines these features with obstacle perception and the current proprioceptive state, while the same estimate is reused for feedforward compensation. As illustrated in Fig.~\ref{flight example}, WAND is deployed onboard for real-time navigation in a cluttered indoor environment under fan-induced airflow.

Our contributions can be summarized as follows:
\begin{itemize}
    \item We propose WAND, a wind-aware RL framework for quadrotor navigation in cluttered environments under wind disturbances. WAND explicitly estimates wind-induced disturbance acceleration from proprioceptive histories and integrates it into the policy through a zero-initialized residual \emph{WindAdapter}, while also reusing it for feedforward compensation during execution.
    
    \item We validate WAND through simulation and real-world flight experiments under diverse wind conditions, demonstrating improved robustness and disturbance-adaptive behaviors compared with baseline methods.
\end{itemize}

\section{RELATED WORK}
\subsection{Hierarchical Navigation in Cluttered Environments}
Autonomous navigation in cluttered environments has been widely studied within hierarchical planning frameworks. Representative methods such as Fast-Planner \cite{fast_planner2019} and EGO-Planner \cite{ego_planner2021} achieve real-time collision-free trajectory generation by exploiting geometric constraints derived from occupancy maps. More recent works \cite{gcopter2022, super2025} further improve computational efficiency and trajectory quality in dense obstacle fields, demonstrating strong performance in structured navigation tasks. However, these planners are primarily designed for geometric feasibility and typically rely on the low-level controller to track the reference trajectory with sufficient accuracy. To improve tracking robustness against external disturbances, prior works have explored disturbance-compensation strategies \cite{cmpcc2021}. Nevertheless, in most hierarchical pipelines, disturbance rejection is primarily handled by the tracking controller. As a result, the planned trajectory may not explicitly account for time-varying aerodynamic effects, and severe wind gusts can drive the quadrotor beyond the intended safety margins, increasing the risk of collision.

\subsection{Learning-Based Navigation}
Learning-based navigation directly learns obstacle-avoidance and goal-reaching behaviors from data. Early learning-based approaches demonstrated sim-to-real navigation in clutter with CAD2RL \cite{cad2rl2016} and privileged learning for high-speed flight in unstructured environments \cite{wild2021}. Building on this, Song \emph{et al.} \cite{learning_agile2023} combine imitation and reinforcement learning through a privileged learning-by-cheating framework that tightly couples perception and control. Xu \emph{et al.} \cite{cloud2025} and Fan \emph{et al.} \cite{e2e_learning2025} design task-relevant LiDAR representations for safe and reactive navigation, while NavRL \cite{navrl2025} addresses dynamic environments with moving obstacles through tailored state representations and adaptive architectures. Yu \emph{et al.} \cite{mav2025} leverage memory-augmented latent representations with varying-speed strategies to enhance navigation performance.

History-conditioned latent adaptation has been applied to quadrotor control by inferring a latent representation of vehicle and environment parameters from recent state--action histories and conditioning the controller on that representation \cite{zhang_quad2023}. In WAND, temporal conditioning is grounded in a supervised estimate of wind-induced disturbance acceleration, thereby making the conditioning signal physically interpretable. Complementary onboard estimators recover wind or external forces from proprioceptive or visual--inertial measurements \cite{DOB2024,VIMO2019,HDVIO2023}, while adaptive learning and control methods improve dynamics prediction or trajectory tracking under wind disturbances \cite{neural_fly2022,piwan2025}. WAND integrates these directions by using the explicit disturbance estimate both to adapt obstacle-aware navigation actions and to provide low-level feedforward compensation.

\begin{figure*}[t] 
    \centering
    \includegraphics[width=\linewidth]{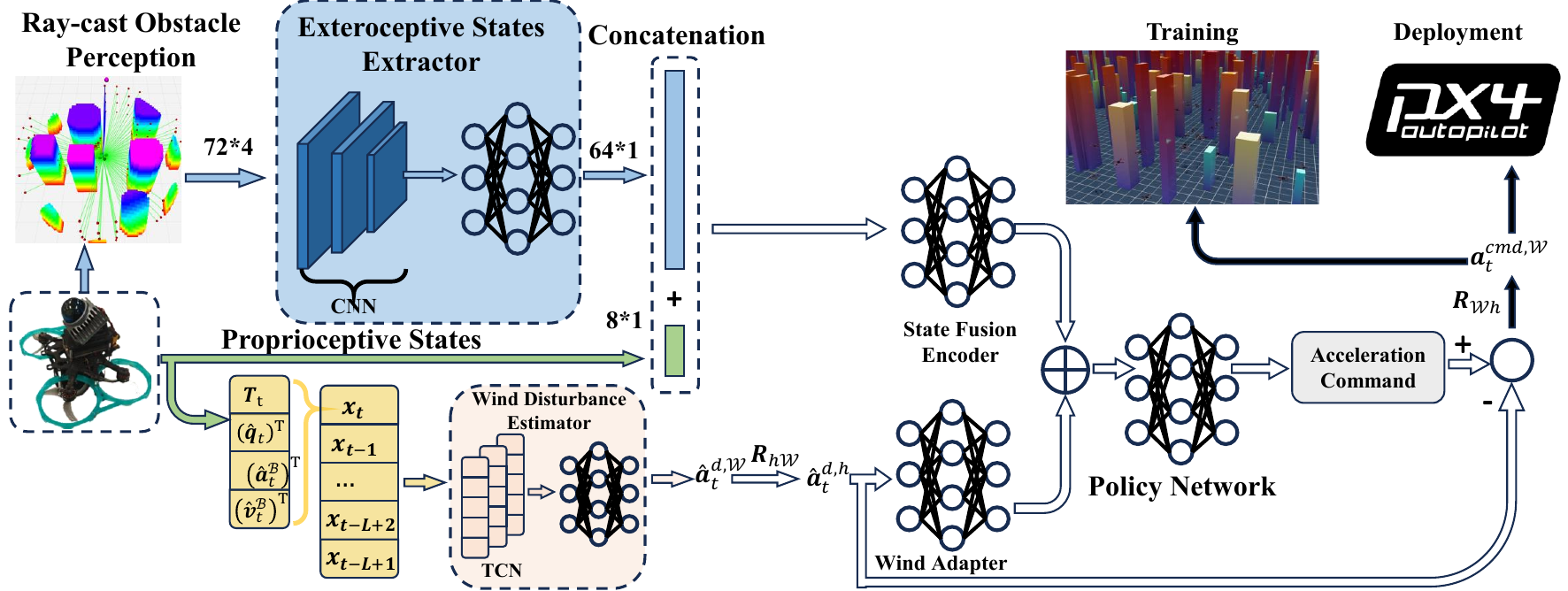}
    \caption{Overview of WAND framework. Ray-cast obstacle perception is first encoded into compact features, while a TCN-based wind disturbance estimator predicts the wind-induced disturbance acceleration from a temporal history of proprioceptive states. The estimated disturbance is encoded by \emph{WindAdapter} and fused with the state feature to form the input to the policy network. The policy outputs a desired acceleration command, which is further combined with the feedforward disturbance compensation term for execution.}
    \label{system-overview}
\end{figure*}

\section{METHODOLOGY}

Fig.~\ref{system-overview} summarizes the disturbance estimator, wind-aware policy, and simulation pipeline of WAND.



\subsection{Wind Disturbance Estimator} \label{wind estimator}
\textbf{Dynamic Model}. Let $\mathcal{W}$ and $\mathcal{B}$ denote the inertial world frame and the quadrotor body frame, respectively. The position, velocity, and acceleration expressed in $\mathcal{W}$ are denoted by $\bm{p}^{\mathcal{W}}$, $\bm{v}^{\mathcal{W}}$, and $\bm{a}^{\mathcal{W}}$. The attitude is represented by the quaternion $\bm{q}$, and $\bm{R}_{\mathcal{W}\mathcal{B}}(\bm{q})\in\mathrm{SO}(3)$ maps vectors from $\mathcal{B}$ to $\mathcal{W}$. Since the estimator targets translational disturbances, the parallel rotor axes are aggregated at the vehicle center of mass and aerodynamic moments are omitted. We define the relative air velocity and rotor-axis direction as
\begin{equation}
\bm{v}_{r}^{\mathcal{W}}=\bm{v}^{\mathcal{W}}-\bm{v}_{w}^{\mathcal{W}},
\qquad
\bm{n}^{\mathcal{W}}=\bm{R}_{\mathcal{W}\mathcal{B}}(\bm{q})\bm{e}_3,
\label{relative_airspeed}
\end{equation}
where $\bm{v}_{w}^{\mathcal{W}}$ is the wind velocity and $\bm{e}_3=[0,0,1]^\mathrm{T}$. The nominal collective-thrust force is
\vspace{-0.2cm}
\begin{equation}
\bm{F}_{T}^{\mathcal{W}}=F_{ct}\bm{n}^{\mathcal{W}},
\qquad
F_{ct}=k_f\sum_{i=1}^{N_r}\omega_i^2,
\label{nominal_thrust}
\end{equation}
where $N_r$ is the number of rotors, $\omega_i$ is the angular speed of rotor $i$, and $k_f$ is the thrust coefficient. Following the first-order rotor model in \cite{drag2018,diff2018}, we define the aerodynamic disturbance as the difference between the total rotor force under relative airflow and the nominal collective thrust:
\vspace{-0.1cm}
\begin{equation}
\begin{aligned}
\bm{F}_{aero}^{\mathcal{W}}(\bm{v}_{w}^{\mathcal{W}})
&=\left[s(\bm{v}_{r}^{\mathcal{W}})-1\right]
F_{ct}\bm{n}^{\mathcal{W}}\\
&\quad-c_d(\bm{\omega})\bm{P}_{n}\bm{v}_{r}^{\mathcal{W}},\\
c_d(\bm{\omega})&=k_r\sum_{i=1}^{N_r}|\omega_i|,\\
\bm{P}_{n}&=\bm{I}-\bm{n}^{\mathcal{W}}
(\bm{n}^{\mathcal{W}})^\mathrm{T}.
\end{aligned}
\label{aerodynamic_force}
\end{equation}
Here, $k_r$ is the rotor-drag coefficient, which determines the magnitude of the aerodynamic force induced by the in-plane relative airflow, and $\bm{P}_{n}$ projects the relative airflow onto the rotor plane. The first-order variation in available thrust caused by axial airflow is represented by the bounded scale factor
\begin{equation}
s(\bm{v}_{r}^{\mathcal{W}})=
\operatorname{sat}_{[s_{\min},s_{\max}]}\!\left(
1-k_s(\bm{v}_{r}^{\mathcal{W}})^\mathrm{T}\bm{n}^{\mathcal{W}}
\right),
\label{thrust_scale}
\end{equation}
where $k_s>0$ is the axial-flow sensitivity coefficient and $0\leq s_{\min}\leq1\leq s_{\max}$ bounds the first-order approximation. The translational dynamics are therefore
\vspace{-0.2cm}
\begin{equation}
m\bm{a}^{\mathcal{W}}
=\bm{F}_{T}^{\mathcal{W}}
+\bm{F}_{aero}^{\mathcal{W}}(\bm{v}_{w}^{\mathcal{W}})
-mg\bm{e}_3.
\label{dynamic}
\end{equation}
WAND estimates the aerodynamic disturbance acceleration
\begin{equation}
\begin{aligned}
\bm{a}^{d,\mathcal{W}}
&:=\frac{\bm{F}_{aero}^{\mathcal{W}}
(\bm{v}_{w}^{\mathcal{W}})}{m}\\
&=\bm{a}^{\mathcal{W}}+g\bm{e}_3
-\frac{\bm{F}_{T}^{\mathcal{W}}}{m}.
\end{aligned}
\label{disturbance_target}
\end{equation}
Thus, $\bm{a}^{d,\mathcal{W}}$ captures the translational acceleration not explained by gravity and nominal collective thrust, including axial thrust variation and rotor drag associated with relative airflow. During Gazebo data collection, the supervisory target is obtained directly as $\bm{a}^{d,\mathcal{W}}_t=\bm{F}^{\mathcal{W}}_{aero,t}/m$, where $\bm{F}^{\mathcal{W}}_{aero,t}$ is the aerodynamic force applied to the vehicle by the simulator at time $t$.

\textbf{Disturbance Estimation}. Although Eq.~\eqref{disturbance_target} provides a physical interpretation of the target, directly evaluating the residual online is sensitive to sensor noise and model mismatch. We instead infer the current disturbance acceleration from a causal history of onboard measurements. At time $t$, the estimator input is
\begin{equation}
\bm{x}_t=\left[T_t,
(\hat{\bm{q}}_t)^\mathrm{T},
(\hat{\bm{a}}_t^{\mathcal{B}})^\mathrm{T},
(\hat{\bm{v}}_t^{\mathcal{B}})^\mathrm{T}\right]^\mathrm{T}
\in\mathbb{R}^{11},
\label{estimator_state}
\end{equation}
where $T_t^{\mathrm{cmd}}$ is the normalized collective-thrust command generated by PX4, and $T_{\mathrm{hover}}$ is the hover-thrust value estimated online by the PX4 Hover Thrust Estimator on the same command scale. Thus, $T_t:=T_t^{\mathrm{cmd}}/T_{\mathrm{hover}}=1$ corresponds to the current PX4 estimate of the nominal hover command. This normalized command is used only as an estimator input. $\hat{\bm{a}}_t^{\mathcal{B}}$ is the body-frame acceleration measured by the IMU, and $\hat{\bm{v}}_t^{\mathcal{B}}$ is the estimated linear velocity.

The TCN maps the causal window $\bm{X}_{t}=[\bm{x}_{t-L+1},\ldots,\bm{x}_{t}]\in\mathbb{R}^{L\times11}$ to the current disturbance acceleration. The TCN consists of three residual causal blocks with kernel size $k{=}3$ and dilations $\{1,2,4\}$, followed by a $1{\times}1$ output convolution. Each causal convolution uses effective left padding $p{=}(k-1)d$, ensuring that the output at time $t$ does not access future measurements:
\begin{equation}
\hat{\bm{a}}_{t}^{d,\mathcal{W}}
=f_{\mathrm{TCN}}(\bm{X}_{t})\in\mathbb{R}^{3}.
\label{disturbance_estimator}
\end{equation}
The estimator captures only the disturbance acting at the vehicle's current location from recent proprioceptive measurements.
We select $L{=}10$ based on a validation sweep over window lengths from 4 to 32 samples. It is the shortest tested window whose validation RMSE is within 1\% of the minimum.

\subsection{Reinforcement Learning Policy}  \label{rl policy}
Since the instantaneous wind disturbance is not directly observed, we formulate wind-aware navigation as a partially observable Markov decision process (POMDP), denoted by $(\mathcal{S},\mathcal{A},\mathcal{O},\mathcal{P},\mathcal{R},\gamma)$. At time $t$, the policy receives an augmented observation $\bm{o}_t$ in which the TCN estimate summarizes the recent proprioceptive history, and selects an action $\bm{u}_t\sim\pi(\cdot\mid\bm{o}_t)$. The objective is
\vspace{-0.2cm}
\begin{equation}
\pi^*=\arg\max_\pi\mathbb{E}\left[\sum_{t=0}^T
\gamma^t R(\bm{s}_t,\bm{u}_t)\right],
\end{equation}
where $\bm{s}_t$ is the latent environment state and $\gamma\in(0,1)$ is the discount factor.

\textbf{Observation}. As shown in Fig. \ref{system-overview}, $\bm{o}_t$ comprises proprioceptive, exteroceptive, and estimated-disturbance information. We express all vector-valued policy inputs in a heading-aligned local frame $h$, whose origin is at the quadrotor, x-axis follows its current horizontal heading, and z-axis is aligned with the world vertical; the y-axis completes the right-handed frame. The policy action is defined in the same frame. The proprioceptive component is
\vspace{-0.2cm}
\begin{equation}
S_{prop,t}=\left[
\left(\frac{\bm p_{g,t}^{h}}{\lVert\bm p_{g,t}^{h}\rVert}\right)^{\mathrm T},
d_{xy,t},d_{z,t},(\bm v_t^{h})^{\mathrm T}
\right]^{\mathrm T},
\end{equation}
where $\bm p_{g,t}^{h}$ is the displacement from the quadrotor to the goal, $d_{xy,t}$ and $d_{z,t}$ are its horizontal distance and signed vertical displacement, respectively, and $\bm v_t^{h}$ is the vehicle velocity.

For exteroception, we use 3D LiDAR range measurements represented by the ray-cast observation of \cite{omnidrones2024}, as illustrated in Fig. \ref{flight example}(b). Rays are uniformly cast along heading-frame unit directions $\hat{\bm e}_{i,j}^{h}$ over $N_h$ horizontal and $N_v$ vertical angles. Let $r_{i,j,t}$ denote the corresponding range measurement. The ray-cast input is $S_{obs,t}=[\tilde r_{i,j,t}]_{N_h\times N_v}$, where $\tilde r_{i,j,t}=3/\operatorname{clip}(r_{i,j,t},0.3,12)-1$; no-hit rays are assigned the upper clipping bound. A lightweight CNN encodes $S_{obs,t}$ into an obstacle feature, which is concatenated with $S_{prop,t}$ and processed by the \emph{State Fusion Encoder} to produce $\bm z_{nav}$.

Before being provided to the policy, the world-frame disturbance estimate is transformed into frame $h$ as $\hat{\bm a}^{d,h}_t=\bm R_{h\mathcal W}\hat{\bm a}^{d,\mathcal W}_t$, where $\bm R_{h\mathcal W}=\bm R_{\mathcal W h}^{\mathrm T}$. The \emph{WindAdapter} encodes $\hat{\bm a}^{d,h}_t$ into a wind-conditioned feature $\bm z_{wind}$ with the same dimensionality as $\bm z_{nav}$. The two features are fused by element-wise addition:
\vspace{-0.2cm}
\begin{equation}
  \bm{z} = \bm{z}_{nav} + \bm{z}_{wind}.
\end{equation}
The output layer of the \emph{WindAdapter} is initialized with zero weights and biases, following the residual initialization principle of ControlNet~\cite{controlnet}. Consequently, $\bm z_{wind}=\bm 0$ and $\bm z=\bm z_{nav}$ at initialization, allowing the wind-conditioned residual to be introduced progressively during joint training.

\textbf{Action}. We optimize the policy using Proximal Policy Optimization (PPO)~\cite{ppo2017}. The actor outputs the desired acceleration $\bm u_t=\bm a_t^{des,h}\in[-3,3]^3\,\mathrm{m/s^2}$ in the heading frame. This bound is applied before feedforward disturbance compensation, after which the command is transformed into the world frame as
\begin{equation}
\bm a_t^{cmd,\mathcal W}
=\bm R_{\mathcal W h}
\left(\bm a_t^{des,h}-\hat{\bm a}_t^{d,h}\right),
\label{executed_command}
\end{equation}
where $\bm R_{\mathcal W h}$ maps vectors from $h$ to $\mathcal W$. The resulting world-frame acceleration setpoint is sent to PX4 through the offboard interface.

\textbf{Reward}. The overall reward $r$ is defined as a weighted sum of five terms: 
\begin{equation}
  r = \lambda_{1} r_{vel} + \lambda_{2} r_{obs} + \lambda_{3} r_{height} + \lambda_{4} r_{wind} + r_{term}.
\end{equation}
The individual reward terms are designed as follows:
\vspace{-0.2cm}
\begin{subequations}
\begin{align}
    &r_{vel} = \frac{\bm{p}^h_{g,t}}{\lVert \bm{p}^h_{g,t} \rVert} \cdot \bm{v}^{h}_t, \\
    &r_{obs} = -\exp\!\left(-2\left(\min\nolimits_{(i,j)} r_{i,j,t} - 0.5\right)\right), \\
    &r_{height} = -\bigl(\min(|p_{c, z} -  p_{s, z}|, |p_{c, z} -  p_{g, z}|)\bigr)^2, \\
    &r_{wind} = -\Bigl[\lVert\hat{\bm{a}}^{d,h}_{t}\rVert \cdot \operatorname{ReLU}\!\bigl(0.3 \lVert\hat{\bm{a}}^{d,h}_{t}\rVert + 0.5 - d_{wind,t}\bigr)\Bigr]^2, \\
    &r_{term} = \begin{cases}
        100, &\text{if Reach Goal} \\
        -50, & \text{if Collision}.
    \end{cases}
\end{align}
\end{subequations}
Here, $r_{vel}$ rewards motion toward the goal, $r_{obs}$ promotes obstacle clearance, and $r_{height}$ discourages excessive vertical deviation; $p_{c,z}$, $p_{s,z}$, and $p_{g,z}$ denote the current, start, and goal heights, respectively. The directional clearance used by $r_{wind}$ is defined as
\vspace{-0.2cm}
\begin{equation}
\begin{aligned}
(i^*,j^*)
&=\mathop{\arg\max}_{(i,j)}
(\hat{\bm e}_{i,j}^{h})^{\mathrm T}
\frac{\hat{\bm a}^{d,h}_t}
{\lVert\hat{\bm a}^{d,h}_t\rVert},\\
d_{wind,t}
&=r_{i^*,j^*,t},\qquad
\lVert\hat{\bm a}^{d,h}_t\rVert>0.1~\mathrm{m/s^2}.
\end{aligned}
\label{wind_direction_clearance}
\end{equation}
For smaller estimated disturbances, $r_{wind}$ is set to zero. Thus, $d_{wind,t}$ is the LiDAR range most closely aligned with the current estimated disturbance direction. The $\operatorname{ReLU}(\cdot)$ term penalizes insufficient directional clearance and encourages a larger local safety margin as the estimated disturbance increases. By explicitly coupling the estimated disturbance with directional obstacle clearance, this term encourages the policy to adapt its safety margin to the disturbance magnitude and direction. We use $\lambda_1=1.0$, $\lambda_2=2.0$, $\lambda_3=8.0$, and $\lambda_4=2.0$ during training.

 \begin{figure}[t] 
    \centering
    \includegraphics[scale=0.26]{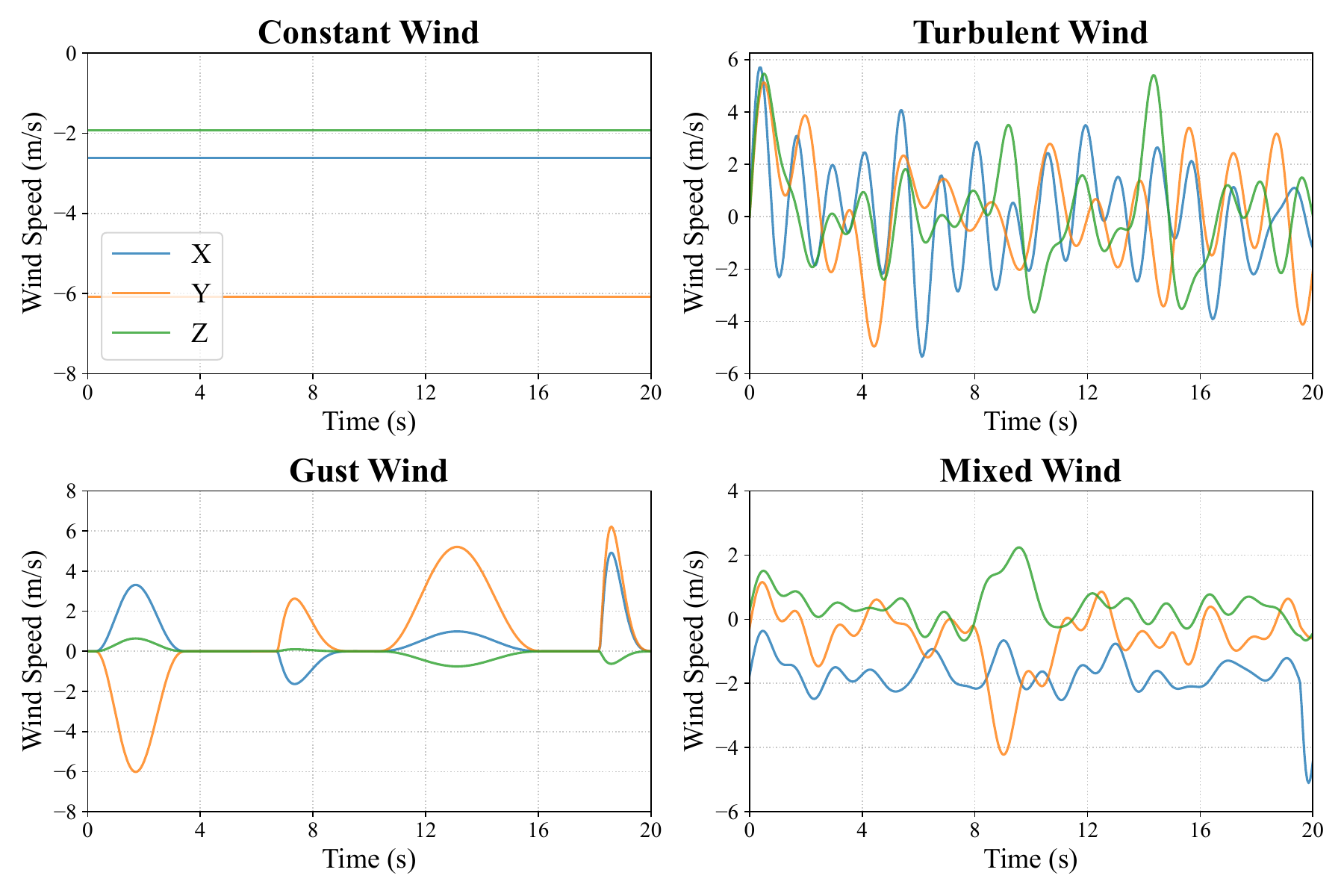}
    \caption{Analytical wind-velocity profiles used in simulation.}
    \label{wind fields}
\end{figure}

\subsection{Simulation and Training Setup}     \label{environment}
\textbf{Wind-Profile Generation}. For standard policy training and simulation evaluation, we define four analytical profiles of the world-frame wind velocity $\bm{v}_{w}^{\mathcal{W}}(t)$ with distinct temporal characteristics, as illustrated in Fig.~\ref{wind fields}. Each standard profile is spatially uniform within the simulated environment and, except for constant wind, varies only over time; its type and parameters are resampled at each episode reset. Thus, these profiles do not model obstacle-induced spatial effects such as wakes or recirculation. We use the scalar amplitude $A$ and the unit direction $\bm{d}^{\mathcal{W}}=[d_x,d_y,d_z]^\mathrm{T}$, where $\|\bm{d}^{\mathcal{W}}\|_2=1$. For turbulent wind, component $i$ has an independently sampled unit direction $\bm d_i^{\mathcal W}$ satisfying $\|\bm d_i^{\mathcal W}\|_2=1$.

\textit{a) Constant Wind}: A steady wind with constant amplitude and direction is defined as
\vspace{-0.1cm}
\begin{equation}
    \bm{v}_{w,\mathrm{const}}^{\mathcal{W}}(t)
    =A\bm{d}^{\mathcal{W}}.
\end{equation}

\textit{b) Turbulent Wind}: Time-varying wind is generated by superimposing $n$ sinusoidal components:
\vspace{-0.1cm}
\begin{equation}
    \bm{v}_{w,\mathrm{turb}}^{\mathcal{W}}(t)
    =\sum_{i=1}^{n}A_i\sin(\Omega_i t+\phi_i)
    \bm d_i^{\mathcal W},
\end{equation}
where $\Omega_i$, $\phi_i$, and $\bm d_i^{\mathcal W}$ denote the angular frequency, initial phase, and direction of component $i$, respectively.

\textit{c) Gust Wind}: Let $\{t_\ell\}$ denote the arrival times of a homogeneous Poisson process with rate $\lambda$, $\tau_\ell=(t-t_\ell)/T_{\mathrm{gust}}$, and $c_{\alpha,\beta}=\max_{0\leq u\leq1}u^{\alpha-1}(1-u)^{\beta-1}$. Each arrival generates a finite-duration beta-shaped gust:
\vspace{-0.1cm}
\begin{equation}
    \bm{v}_{w,\mathrm{gust},\ell}^{\mathcal{W}}(t)
    =\frac{A\bm{d}^{\mathcal{W}}}{c_{\alpha,\beta}}
    \tau_\ell^{\alpha-1}(1-\tau_\ell)^{\beta-1}.
\end{equation}
This expression applies for $0\leq\tau_\ell\leq1$ and is zero otherwise. The normalization makes $A$ the peak gust speed; $T_{\mathrm{gust}}$ is the gust duration, while $\alpha$ and $\beta$ control its rise--decay shape. Simultaneously active pulses are summed.

\textit{d) Mixed Wind}: The three components are combined as
\vspace{-0.1cm}
\begin{equation}
    \bm{v}_{w,\mathrm{mix}}^{\mathcal{W}}(t)
    =\sum_{k\in\mathcal{K}}\varepsilon_k
    \bm{v}_{w,k}^{\mathcal{W}}(t),
\end{equation}
where $\mathcal{K}=\{\mathrm{const},\mathrm{turb},\mathrm{gust}\}$ and
$\varepsilon_k=u_k/\sum_{j\in\mathcal{K}}u_j$ with
$u_k\sim\mathcal{U}(0,1)$.

For policy training, $V_{\max}=15\,\mathrm{m/s}$, $n=8$, and $\lambda=0.2\,\mathrm{s}^{-1}$. Constant- and gust-wind amplitudes are sampled from $\mathcal{U}(0.5V_{\max},V_{\max})$, while $A_i\sim\mathcal{U}(0.5V_{\max}/n,V_{\max}/n)$, $\Omega_i\sim\mathcal{U}(0.5,1.5)\,\mathrm{rad/s}$, and $\phi_i\sim\mathcal{U}(0,2\pi)$. For gusts, $T_{\mathrm{gust}}\sim\mathcal{U}(2,8)\,\mathrm{s}$, $\alpha\sim\mathcal{U}(3,5)$, and $\beta\sim\mathcal{U}(5,8)$. All direction vectors, including $\bm d^{\mathcal W}$ and $\bm d_i^{\mathcal W}$, are randomly sampled and normalized to unit length.

\textbf{Aerodynamic Parameters}. For the model in Eqs.~\eqref{aerodynamic_force} and~\eqref{thrust_scale}, the effective rotor-drag coefficient is resampled at each episode as $k_r\sim\mathcal{U}(6,8)\times10^{-5}\,\mathrm{kg}$. We set $k_s=0.04\,\mathrm{s/m}$ and $[s_{\min},s_{\max}]=[0,2]$.

\textbf{Simulation Environments}. Policy training is conducted in Isaac Sim using 1200 parallel environments, each containing 500 obstacles within a $40\,\mathrm{m}\times40\,\mathrm{m}$ area. Wind profiles are randomized between episodes as described above, and the ray-cast observation uses $N_h=72$ and $N_v=4$. The Gazebo environment used for estimator-data collection and sim-to-sim evaluation consists of a $20\,\mathrm{m}\times20\,\mathrm{m}$ map populated with 90 static obstacles.

\textbf{Estimator Training and Transfer}. We collect 90{,}000 labeled samples at 30~Hz in Gazebo under the constant, turbulent, and gust profiles, with wind-speed levels ranging from $0$ to $7.5\,\mathrm{m/s}$. The supervisory target is the simulator-applied aerodynamic force divided by the vehicle mass, as defined in Eq.~\eqref{disturbance_target}. Complete episodes are divided into disjoint training, validation, and test sets in a 60/20/20 ratio before temporal-window extraction. The TCN is trained with MSE loss using Adam for up to 60 epochs, with a learning rate of $10^{-4}$ and a batch size of 64. A separate mixed-wind dataset is generated exclusively for held-out evaluation and is not used for training, validation, or model selection.

Gazebo and Isaac Sim implement the same aerodynamic-force model and parameterization in Eqs.~\eqref{aerodynamic_force}--\eqref{thrust_scale}, together with the same coordinate conventions and estimator input definition. The Gazebo-trained estimator is therefore transferred to Isaac Sim without retraining and remains frozen throughout PPO optimization. Both policy conditioning and feedforward compensation use its predictions rather than simulator ground truth. For real-world deployment, the policy is transferred without further training, whereas the estimator is fine-tuned using real-flight data as described in Sec.~\ref{real-world experiment}.

\textbf{Policy Training}. The policy is optimized for 400 million environment steps. The navigation encoder, \emph{WindAdapter}, actor, and critic are jointly updated using PPO. The \emph{WindAdapter} output layer is zero initialized, and the remaining policy layers use standard random initialization. We use Adam with a learning rate of $5\times10^{-4}$ and a discount factor of 0.99. Training takes approximately 3 hours on an NVIDIA RTX 4090 GPU.


\section{EXPERIMENTS AND RESULTS}

\begin{figure}[t] 
    \centering
    \includegraphics[width=\linewidth]{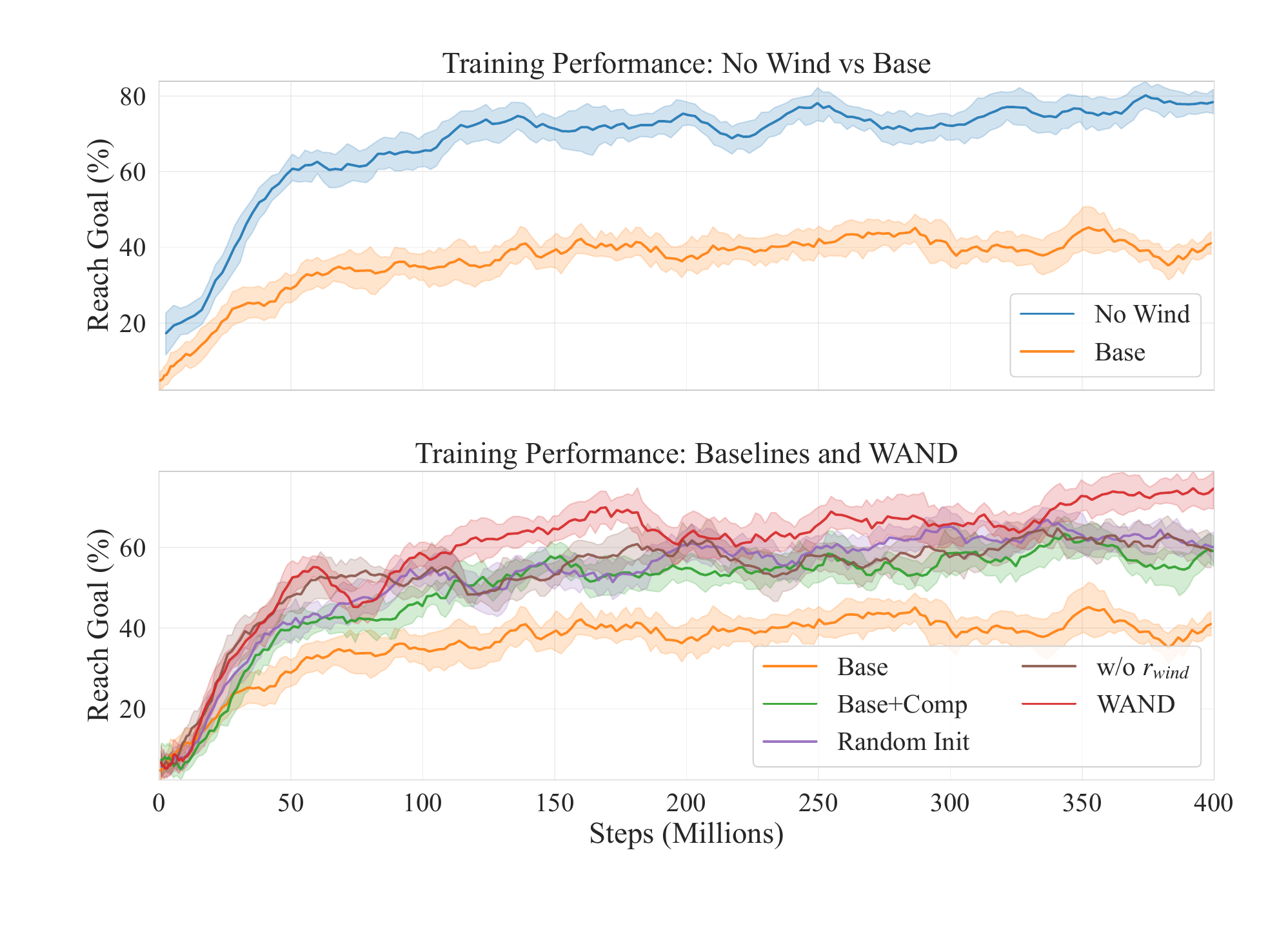}
    \caption{Reach-goal rates during PPO training: (a) navigation-only
    comparison and (b) WAND ablations. Curves show the mean over five
    independent random seeds, and shaded regions indicate 95\% confidence
    intervals across seeds.}
    \label{training}
\end{figure}

\begin{figure*}[t]
    \centering
    \includegraphics[scale=0.48]{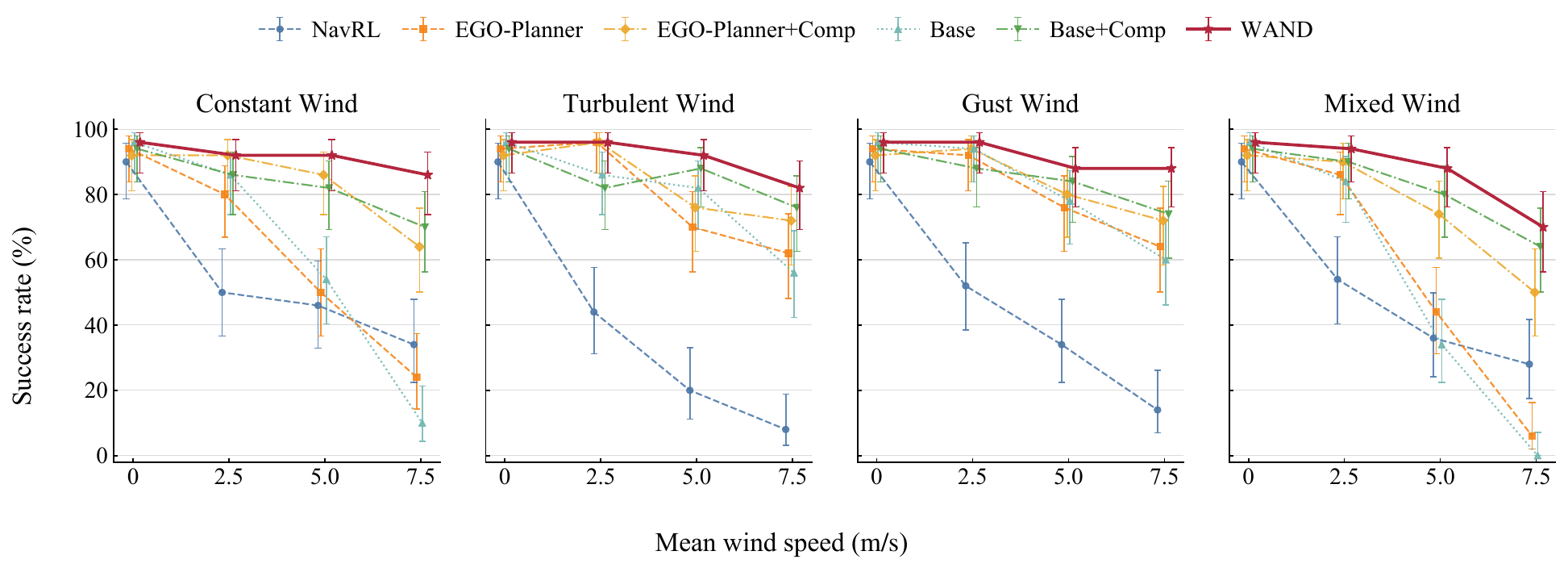}
    \caption{Navigation success rates under constant, turbulent, gust, and mixed wind. Markers show success rates over 50 trials; error bars indicate 95\% Wilson confidence intervals. Markers are slightly offset horizontally for clarity.}
    \label{fig:success_ci}
\end{figure*}

\subsection{RL Training Results}     \label{rl training}
To isolate the contributions of disturbance compensation, policy conditioning, and the interaction between zero initialization and reward shaping, we compare six configurations. ``No Wind'' and ``Base'' share the navigation-only policy and are trained without and with wind disturbances, respectively, while ``Base+Comp'' adds TCN-based feedforward compensation to Base. The remaining variants use the \emph{WindAdapter}: ``w/o $r_{wind}$'' retains zero initialization but uses the same navigation reward as Base+Comp, ``Random Init'' retains $r_{wind}$ but randomly initializes the \emph{WindAdapter} output layer, and ``WAND'' combines zero initialization with $r_{wind}$. All wind-disturbed configurations share the same environment randomization and PPO settings.

Fig.~\ref{training}(a) shows that wind disturbances substantially reduce the reach-goal rate of the navigation-only policy. In Fig.~\ref{training}(b), Base+Comp improves over Base, confirming the benefit of disturbance feedforward compensation. In contrast, neither the zero-initialized \emph{WindAdapter} without $r_{wind}$ nor $r_{wind}$ with random initialization provides a marked improvement over Base+Comp. Combining zero initialization and $r_{wind}$ in WAND produces a clear increase in the late-stage reach-goal rate. This result suggests that the two designs are complementary: zero initialization facilitates the progressive integration of disturbance-conditioned features, while $r_{wind}$ provides the corresponding navigation-level learning signal.

\subsection{Simulation Results}     \label{simulation experiment}

We conduct sim-to-sim evaluations in Gazebo using the environment described in Sec.~\ref{environment}. We first compare EKF, LSTM, and TCN using the RMSE of the estimated disturbance acceleration. Results for constant, turbulent, and gust wind are computed on the corresponding episode-level test sets, while mixed wind is evaluated on the separately generated held-out composite set. As shown in Table~\ref{tab:pre_rmse}, TCN achieves the lowest RMSE under all four conditions, including the held-out mixed-wind condition.

\begin{table}[t]
    \centering
    \caption{RMSE($m/s^2$) IN DIFFERENT WIND TYPES}
    \label{tab:pre_rmse}
    \begin{tabular*}{\columnwidth}{@{\extracolsep{\fill}}lcccc@{}}
    \toprule
    \textbf{Method} & \textbf{Constant} & \textbf{Turbulent} & \textbf{Gust} & \textbf{Mixed} \\
    \midrule
    EKF  & 0.3204 & 0.4620 & 0.4203 & 0.4474 \\
    LSTM & 0.1166 & 0.2405 & 0.2059 & 0.2525 \\
    TCN  & \textbf{0.0777} & \textbf{0.1876} & \textbf{0.1159} & \textbf{0.1970} \\
    \bottomrule
    \end{tabular*}
\end{table}

We then assess the navigation framework. We evaluate several mean wind speeds for each disturbance type. For a fair comparison, all methods are tested under the same environment settings and wind conditions. For each setting, we conduct 50 randomized trials. We compare our method with two prior navigation approaches, NavRL \cite{navrl2025} and EGO-Planner \cite{ego_planner2021}, both of which are widely used open-source methods. In addition, we include an ``EGO-Planner+Comp'' variant, which augments EGO-Planner with the same feedforward compensation term predicted by TCN. We also include the ``Base'' and ``Base+Comp'' ablations to disentangle the effects of feedforward compensation and wind-aware policy adaptation.

As shown in Fig.~\ref{fig:success_ci}, the performance of NavRL, EGO-Planner, and Base generally degrades as the wind intensity increases, while feedforward compensation generally mitigates this degradation. WAND achieves the highest or tied-highest observed success rate in all 13 settings. Compared with Base+Comp, WAND obtains a higher observed success rate in all 12 wind-disturbed settings, with an average improvement of 8.3 percentage points. These consistent gains across wind profiles and intensities support the benefit of combining feedforward disturbance compensation with wind-conditioned policy adaptation.

To assess how the navigation policy responds to wind direction, we further compare WAND with Base+Comp in a fixed obstacle layout under opposite crosswinds. The start and goal are fixed at $\bm p_s^{\mathcal W}=[0,0,1]^{\mathrm T}\,\mathrm m$ and $\bm p_g^{\mathcal W}=[0,15,1]^{\mathrm T}\,\mathrm m$, respectively. A constant $7.5\,\mathrm{m/s}$ crosswind is applied along either $+\mathcal W x$ or $-\mathcal W x$, while the map and wind-exposure region remain unchanged. Each method is evaluated in 20 trials per wind direction.

\begin{figure}[t]
    \centering
    \begin{subfigure}[t]{0.4\textwidth}
        \centering
        \includegraphics[width=\linewidth]{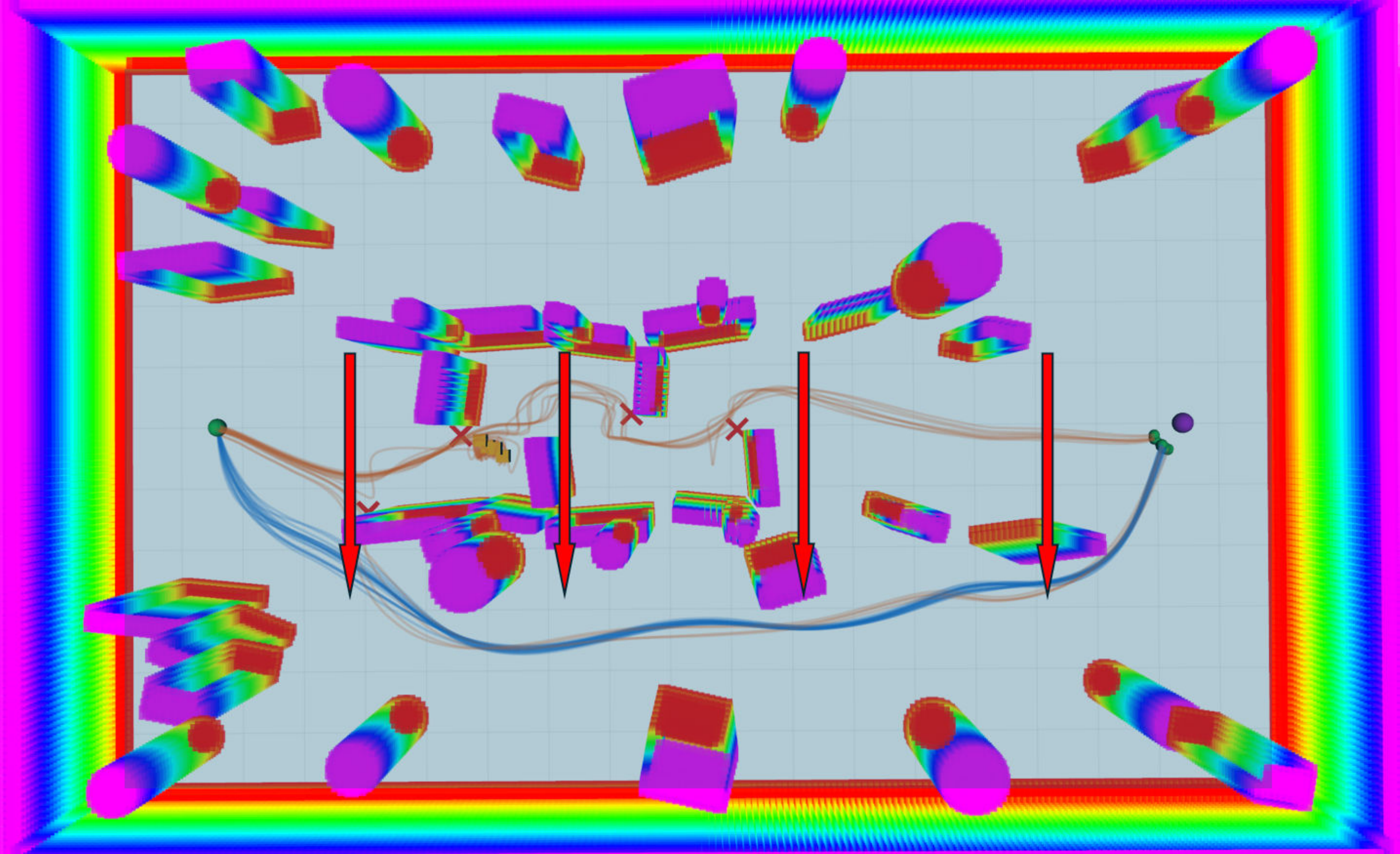}
        \caption{Crosswind along $+\mathcal W x$.}
    \end{subfigure}
    \vspace{1mm}
    \begin{subfigure}[t]{0.4\textwidth}
        \centering
        \includegraphics[width=\linewidth]{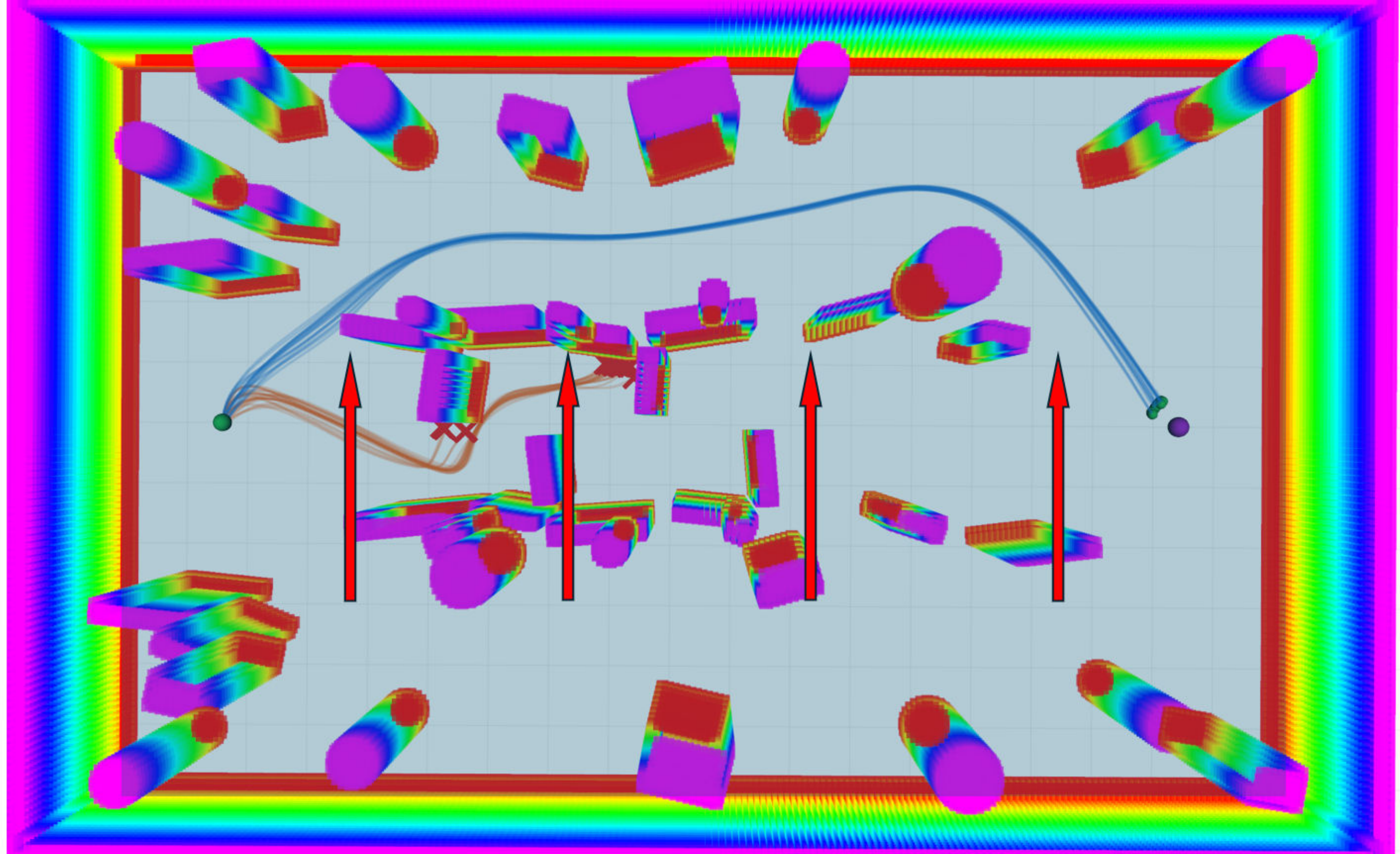}
        \caption{Crosswind along $-\mathcal W x$.}
    \end{subfigure}
    \caption{Fixed-scene trajectories under opposite
    $7.5\,\mathrm{m/s}$ crosswinds. The translucent volume and arrows
    indicate the prescribed wind region and direction. Each panel
    overlays 20 trials per method; WAND and Base+Comp are shown in blue
    and orange, while red crosses and yellow squares mark collisions and
    timeouts.}
    \label{fig:fixed_scene}
\end{figure}

\begin{table}[t]
    \centering
    \caption{FIXED-SCENE RESULTS UNDER OPPOSITE CROSSWINDS.}
    \label{tab:fixed_scene}
    \begin{tabular*}{\columnwidth}
        {@{\extracolsep{\fill}}llcc@{}}
        \toprule
        \textbf{Wind} & \textbf{Method} &
        \textbf{S/C/T (\%)} & $\bm{d_{\min}}$ \textbf{(m)} \\
        \midrule
        $+\mathcal W x$ & Base+Comp & 60/20/20 & $0.086\pm0.089$ \\
                       & WAND      & \textbf{100/0/0} &
                         $\mathbf{0.376\pm0.022}$ \\
        \midrule
        $-\mathcal W x$ & Base+Comp & 0/100/0 & -- \\
                       & WAND      & \textbf{100/0/0} &
                         $\mathbf{0.449\pm0.107}$ \\
        \bottomrule
    \end{tabular*}
    \vspace{1mm}
    \parbox{\columnwidth}{\footnotesize
    S/C/T denotes success, collision, and timeout rates. Each condition
    contains 20 trials. Minimum obstacle clearance is computed over
    successful trials and reported as the mean $\pm$ standard deviation
    of the per-trial minima; -- indicates that no trial succeeded.}
\end{table}

Fig.~\ref{fig:fixed_scene} shows that reversing the crosswind shifts WAND's trajectories to the opposite side of the central obstacle cluster. WAND succeeds in all trials under both directions, whereas Base+Comp achieves success rates of 60\% and 0\%, respectively (Table~\ref{tab:fixed_scene}). Under $+\mathcal W x$, WAND also increases the mean minimum clearance of successful trials by $0.290\,\mathrm m$, demonstrating the benefit of disturbance-conditioned navigation beyond feedforward compensation.

\begin{figure*}[t]
    \centering
    \includegraphics[width=0.9\textwidth]{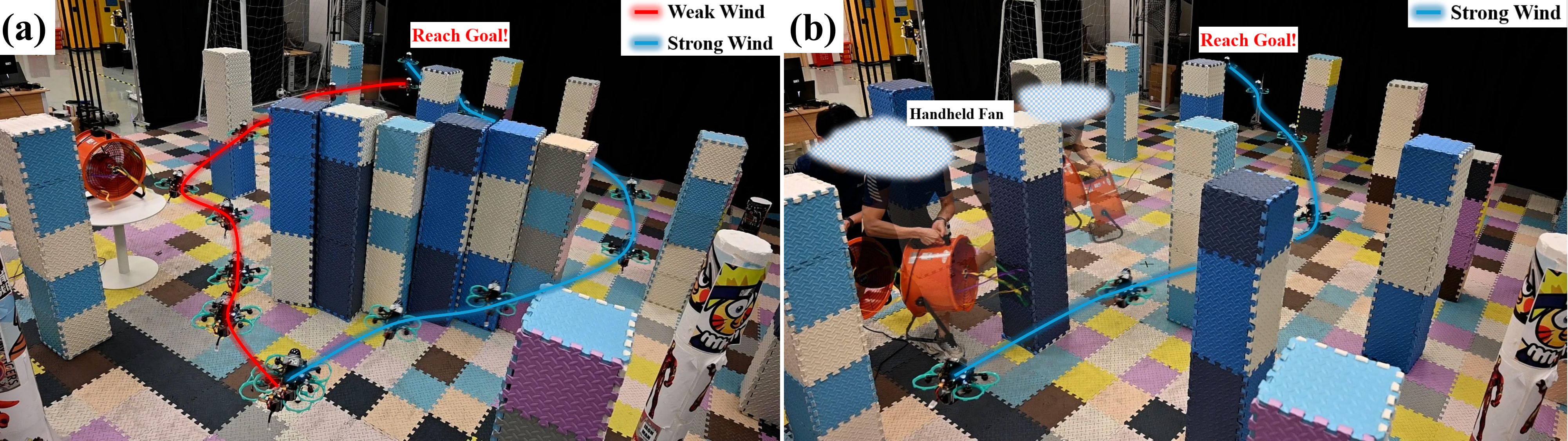}
    \caption{Real-world trajectories under (a) weak and strong fixed-fan wind and (b) strong handheld-fan wind.}
    \label{fig:real_scenes}
\end{figure*}

\subsection{Real-World Experiments}     \label{real-world experiment}
To evaluate onboard deployment, we conduct real-world navigation experiments in cluttered environments under fan-induced aerodynamic disturbances, as shown in Fig.~\ref{fig:real_scenes}. The quadrotor carries an NVIDIA Jetson Orin NX and a Livox Mid-360 3D LiDAR. The LiDAR stream is processed by Super-LIO \cite{super-lio} for real-time state estimation and is also converted into obstacle observations for the policy. The TCN estimator and PPO navigation module contain 196{,}547 and 146{,}151 parameters, respectively, for 342{,}698 parameters in total. Both modules run onboard at 30~Hz, with an average combined inference time of $11.76\,\mathrm{ms}$. The desired acceleration commands are sent to the PX4 flight controller through its control interface.

For real-world deployment, the navigation policy is transferred directly from simulation without further training, whereas the Gazebo-trained disturbance estimator is fine-tuned using approximately 10~min of real-flight data, corresponding to about 18{,}000 samples at 30~Hz. These data are collected during separate manually piloted flights and exclude all autonomous navigation-evaluation trials. The supervisory target is computed according to Eq.~\eqref{disturbance_target} using the world-frame acceleration and attitude provided by motion capture together with the corresponding collective-thrust command. The acceleration output is used directly without numerical differentiation or additional filtering, and no explicit delay compensation is applied. Motion capture is used only for offline label generation and is not involved in online navigation. On held-out real-flight sequences from Scenes~1 and~2, the fine-tuned TCN achieves RMSEs of $0.2179$ and $0.2442\,\mathrm{m/s^2}$, respectively.

\begin{table}[t]
    \centering
    \caption{REAL-WORLD NAVIGATION RESULTS (SUCCESSES/TRIALS)}
    \label{tab:real_scene}
    \begin{tabular*}{\columnwidth}{@{\extracolsep{\fill}}lccc@{}}
    \toprule
    \textbf{Method} & \textbf{Scene 1} & \textbf{Scene 2}
    & \textbf{Total} \\
    \midrule
    EGO-Planner & 4/10 & 6/10 & 10/20 \\
    NavRL       & 5/10 & 6/10 & 11/20 \\
    Base+Comp   & 7/10 & 6/10 & 13/20 \\
    \textbf{WAND} & \textbf{10/10} & \textbf{8/10}
    & \textbf{18/20} \\
    \bottomrule
    \end{tabular*}
\end{table}

Under fan-induced airflow with locally measured speeds up to approximately
$7.2\,\mathrm{m/s}$, each method is evaluated in ten trials per scene
(20 total). WAND succeeds in 18 trials, versus 13 for Base+Comp, 10 for
EGO-Planner, and 11 for NavRL (Table~\ref{tab:real_scene}). Under stronger
fixed-fan wind, WAND shifts from the direct route toward a more open
passage [Fig.~\ref{fig:real_scenes}(a)], while
Fig.~\ref{fig:real_scenes}(b) shows navigation under a handheld-fan
disturbance, providing system-level evidence of disturbance-conditioned
behavior in the evaluated indoor scenes.


\section{CONCLUSIONS}
This paper presented WAND, a wind-aware reinforcement learning framework for quadrotor navigation in cluttered environments under time-varying aerodynamic disturbances. By combining a navigation policy with a wind disturbance estimator and the proposed \emph{WindAdapter} module, WAND enables the quadrotor to adapt its motion strategy according to the wind condition. Experiments in simulation and controlled indoor fan tests demonstrated improved navigation robustness. Systematic evaluation under spatially varying wind fields remains future work.

\bibliographystyle{IEEEtran}
\bibliography{references}

\end{document}